\documentclass[conference]{IEEEtran}
\usepackage[utf8]{inputenc}
\usepackage[T1]{fontenc}
\usepackage{microtype}

\usepackage{todonotes}

\usepackage[keeplastbox]{flushend}

\usepackage{graphicx}

 \usepackage{csquotes}

\usepackage{hyphenat}

\usepackage[capitalize,noabbrev]{cleveref}
\crefname{lstlisting}{listing}{listings}
\Crefname{lstlisting}{Listing}{Listings}
\crefname{listing}{listing}{listings}
\Crefname{listing}{Listing}{Listings}
\crefname{algorithm}{algorithm}{algorithms}
\Crefname{algorithm}{Algorithm}{Algorithms}

\usepackage{booktabs}
\usepackage{url}
\usepackage{microtype}

\RequirePackage{xstring}
\RequirePackage{xparse}
\RequirePackage[]{acro}
\NewDocumentCommand\acrodef{mO{#1}mG{}}{\DeclareAcronym{#1}{short={#2}, long={#3}, #4}}
\acrodef{CPS}{Cyber-Physical System}
\acrodef{MAPE}{\textit{Monitor, Analyze, Plan, Execute}}
\acrodef{MAB-EX}{\textit{Monitor, Analyze, Build, Explain}}
\acrodef{V2X}{\textit{Vehicle\hyp{}to\hyp{}X}}
\acrodef{LSC}{\textit{Live Sequence Charts}}
\acrodef{SML}{\textit{Scenario Modeling Language}}

\newcommand{\absatz}[1]{~\\\textbf{#1:}}

\definecolor{darkgreen}{rgb}{0.0,0.4,0.0} 
\definecolor{darkred}{rgb}{0.6,0.1,0.1}
\definecolor{lightgray}{gray}{.98}
\definecolor{medgray}{gray}{.70}
\definecolor{darkgray}{gray}{.40}
\definecolor{lightviolet}{rgb}{0.7,0,0.7} 
\definecolor{darkviolet}{rgb}{0.5,0.1,0.5}
\definecolor{darkredviolet}{rgb}{0.6,0.1,0.4}
\definecolor{limegreen}{rgb}{0.2,0.7,0.2}
\definecolor{navyblue}{RGB}{0,0,128}
\definecolor{aquamarine}{RGB}{102,205,170}

\definecolor{strictRED}{RGB}{184,0,0}
\definecolor{specificationTURQUOISE}{RGB}{0,128,153}
\definecolor{assumptionGREEN}{RGB}{0,128,0}
\definecolor{interruptBLUE}{RGB}{0,0,128}
\definecolor{committedORCHID}{RGB}{54,22,89}
\definecolor{urgentORCHID}{RGB}{74,28,109}
\definecolor{requestedORCHID}{RGB}{104,34,139}
\definecolor{eventuallyORCHID}{RGB}{154,50,205}

\usepackage{courier}

\usepackage{moresize}
\usepackage{listings}

\usepackage{paralist}

\lstdefinelanguage{SMLX}
{
	basicstyle=\ssmall\ttfamily, %
	frame=single, 
	framextopmargin=0pt,
	framexbottommargin=0pt,
	framexleftmargin=0pt,
	xleftmargin=16pt,
	xrightmargin=3pt,
	morekeywords=[1]{system, domain, scenario, bind, to, 
		message, non, spontaneous, events, specification, 
		alternative, if, collaboration, role, with, dynamic, 
		bindings, or, and, null, define, as, 
		constraints, import, static, parameter, ranges, var, EInt, 
		controllable},
	morekeywords=[2]{strict},
	morekeywords=[3]{forbidden, violation},
	morekeywords=[4]{interrupt},
	morekeywords=[5]{guarantee},
	morekeywords=[6]{assumption}, 
	morekeywords=[7]{committed}, 
	morekeywords=[8]{urgent},
	morekeywords=[9]{requested},
	morekeywords=[10]{eventually},
	keywordstyle=[1]\color{darkviolet}\textbf,
	keywordstyle=[2]\color{strictRED}\textit,
	keywordstyle=[3]\color{strictRED}\textit,
	keywordstyle=[4]\color{interruptBLUE}\textit,
	keywordstyle=[5]\color{specificationTURQUOISE}\textbf,
	keywordstyle=[6]\color{assumptionGREEN}\textbf,
	keywordstyle=[7]\color{committedORCHID}\textit,
	keywordstyle=[8]\color{urgentORCHID}\textit,
	keywordstyle=[9]\color{requestedORCHID}\textit,
	keywordstyle=[10]\color{eventuallyORCHID}\textit,
	sensitive=false,
	morecomment=[l][\color{darkgreen}\textit]{//},
	morecomment=[s][\color{darkgreen}\textit]{/*}{*/}, 
	morestring=[b][\color{blue}]",
	tabsize=1,
	moredelim = [s][\color{specificationTURQUOISE}\textbf]{guarantee}{scenario},
	moredelim = [s][\color{assumptionGREEN}\textbf]{assumption}{scenario},
	backgroundcolor=\color{lightgray}
}

\newcommand{\emodel}{explanation model}
\newcommand{\tmodel}{recipient model}

\lstdefinestyle{SMLXStyle} {language=SMLX}

\makeatletter
\def\ps@IEEEtitlepagestyle{%
  \def\@oddfoot{\mycopyrightnotice}%
  \def\@evenfoot{}%
}
\def\mycopyrightnotice{%
  {\begin{minipage}{\textwidth}
  \footnotesize \copyright 2019 IEEE. Personal use of this material is permitted. Permission from IEEE must be obtained for all other uses, in any current or future media, including reprinting\slash republishing this material for advertising or promotional purposes, creating new collective works, for resale or redistribution to servers or lists, or reuse of any copyrighted component of this work in other works.
  \end{minipage}
  }
  \gdef\mycopyrightnotice{}
}

\title{Towards Self-Explainable Cyber-Physical Systems}

\author{%
\IEEEauthorblockN{%
Mathias~Blumreiter\IEEEauthorrefmark{1}%
,
Joel~Greenyer\IEEEauthorrefmark{2}%
,
Francisco~Javier~Chiyah~Garcia\IEEEauthorrefmark{3}%
,
Verena~Kl{\"o}s\IEEEauthorrefmark{5}%
, \\
Maike~Schwammberger\IEEEauthorrefmark{6}%
,
Christoph~Sommer\IEEEauthorrefmark{7}%
,
Andreas~Vogelsang\IEEEauthorrefmark{8}%
~and
Andreas~Wortmann\IEEEauthorrefmark{9}%
}%
\IEEEauthorblockA{\IEEEauthorrefmark{1}Institute for Software Systems, Hamburg University of Technology, Germany}%
\IEEEauthorblockA{\IEEEauthorrefmark{2}Software Engineering Group, Leibniz Universit{\"a}t Hannover, Germany}%
\IEEEauthorblockA{\IEEEauthorrefmark{3}Heriot-Watt University, United Kingdom}%
\IEEEauthorblockA{\IEEEauthorrefmark{5}Software and Embedded Systems Engineering, Technische Universit{\"a}t Berlin, Germany}%
\IEEEauthorblockA{\IEEEauthorrefmark{6}Department of Computing Science, University of Oldenburg, Germany}%
\IEEEauthorblockA{\IEEEauthorrefmark{7}Heinz Nixdorf Institute and Dept.\ of Computer Science, Paderborn University, Germany}%
\IEEEauthorblockA{\IEEEauthorrefmark{8}Automated Systems Engineering Technologies, Technische Universit{\"a}t Berlin, Germany}%
\IEEEauthorblockA{\IEEEauthorrefmark{9}Software Engineering, RWTH Aachen University, Germany}%
\texttt{mathias.blumreiter@tuhh.de}%
,
\texttt{greenyer@inf.uni-hannover.de}%
,
\texttt{fjc3@hw.ac.uk}%
,\\
\texttt{verena.kloes@tu-berlin.de}%
,
\texttt{schwammberger@informatik.uni-oldenburg.de}%
,\\
\texttt{sommer@ccs-labs.org}%
,
\texttt{andreas.vogelsang@tu-berlin.de}%
,
\texttt{wortmann@se-rwth.de}%
}

\begin{document}

\maketitle

\begin{abstract}\nohyphens{%
With the increasing complexity of \acl{CPS}s, their behavior and decisions become increasingly difficult to understand and comprehend for users and other stakeholders. 
Our vision is to build self-explainable systems that can, at run-time, answer questions about the system's past, current, and future behavior.
%
As hitherto no design methodology or reference framework exists for building such systems, we propose the \ac{MAB-EX} framework for building self-explainable systems that leverage requirements- and explainability models at run-time. The basic idea of \ac{MAB-EX} is to first \emph{\textbf{M}onitor} and \emph{\textbf{A}nalyze} a certain behavior of a system, then Build an explanation from explanation models and convey this EXplanation in a suitable way to a stakeholder. We also take into account that new explanations can be learned, by updating the explanation models, should new and yet un-explainable behavior be detected by the system. 
}
\end{abstract}


\acresetall

\section{Motivation}


The complexity of \ac{CPS} is constantly increasing because they control more and more complex processes in the physical world, possibly with multiple users, changing contexts, and changing environmental conditions.
Hence, their software is distributed, concurrent, and combines discrete and continuous aspects.
Due to this complexity, it becomes increasingly difficult for system- and software engineers, but also users, auditors, and other stakeholders, to comprehend the behavior of a system.
Thus, it will be increasingly relevant for future \acs{CPS} to explain their behavior to 
their stakeholders. This is essential to improve the trust and understanding between the user and the system \cite{Lim2009}, 
to enhance collaboration,
and 
to
increase confidence \cite{LeBras18}.
Our vision is to 
enable the development of 
\emph{self-explainable systems} that can -- at run-time -- answer questions about 
their 
past, current, and future behavior, 
e.g., why a certain action was taken, what goals the system tries to achieve and how, etc.

An example for an ambiguous action that might need explanation could be that a user in an autonomous car wishes to know an answer to the following question: \emph{\enquote{Why are we leaving the highway?}}.
Here, the observed behavior is \enquote{leaving the highway}.
However, there could be several explanations for the behavior, e.g., \emph{\enquote{We are leaving the highway ...}}
\begin{itemize}
    \item \emph{\enquote{... because there is a traffic jam ahead}}; or
    \item \emph{\enquote{... because we reached our travel destination}}; or
    \item \emph{\enquote{... because we need to drive to a gas station}}.
\end{itemize}


Adding such self-explainability capabilities, however, is difficult.
Self-explainability requires that the system has some understanding (i.e., a model) of itself, its environment, the requirements that it shall satisfy, and more: an understanding of the stakeholder that requires an explanation, and mechanisms that can reflect on the current behavior and provide hindsight and foresight.
To date, there is no requirements engineering or design methodology for building such systems, and there is no reference framework for building self-explainable systems.


In this paper, we propose such a reference framework for building self-explainable systems which bases on the \ac{MAPE} loop for self-adaptive systems from IBM~\cite{ibm2005architectural}. 
The \ac{MAPE} loop proposes to continuously \emph{monitor} relevant system and environment data, and, based on this, \emph{analyze} whether an adaptation is necessary to satisfy system goals/ improve the performance. According to the analysis results, the system \emph{plans} and \emph{executes} a suitable adaptation. As we need similar self-reflection capabilities for a self-explaining system, we adapt this feedback loop to our needs. 
We demonstrate the applicability of our approach by sketching realizations in an example use case of a \ac{V2X} driver assistance system, which is prototypical for cooperative mobile systems in smart cities~\cite{sommer2014vehicular}.

We introduce details on our \ac{MAB-EX} framework and 
how
it adapts the \ac{MAPE} loop in \cref{sec:mabex-loop}.
Afterwards, we 
illustrate
its application to our use case in \cref{sec:case-study}.
We discuss challenges yet to face and potential extensions of our framework in \cref{sec:discussion}.
For related work on explainability of \acs{CPS} see the following \cref{sec:related-work}.

\section{Explainability in Software-Intensive \acs{CPS} --\\An Overview}\label{sec:related-work}
Explainability has gained attention due to research projects on \textit{Explainable AI}. Whereas these projects focus on explaining machine learning results, many \acs{CPS} make context-dependent decisions that are not based on ML. To explain these decisions, some approaches focus on \textit{explainable planning}: In~\cite{fan2018generating}, Assumption-based Argumentation is used to model planning problems and to generate explanations for planning solutions as well as for invalid plans. \cite{zhao2019interactive} explicitly focus on \acs{CPS}. This work-in-progress aims at providing interactive explanations based on Why and Why-Not questions from end-users about specific behaviors of the system. Answers are provided in form of contrastive explanations. Explanations contain the consequences or properties of choices, and how the choices affect goals and objectives
of the system. In~\cite{sukkerd2018towards}, verbal explanations of multi-objective probabilistic planning are automatically generated. They also use contrastive justification as explanation for why a generated behavior is preferred to other alternatives.
In contrast to our work, these approaches focus on how to generate explanations and do neither provide a framework for identifying situations that need to be explained nor provide automatic customization to users and operation contexts.

In~\cite{drechsler2018}, the authors sketch first steps towards a conceptual framework for self-explaining CPS. Similar to our approach, they propose to add a layer for self-explanation that includes an abstract model of the system, and they propose to adjust the granularity of explanations for different user groups. In contrast to our work, they propose to construct cause-effect chains for observable actions using the abstract model. Users can access these chains to understand the cause of actions. 

In~\cite{Wuest19}, a feedback loop approach is used to identify situations where it is valuable to ask a user for feedback about system behavior. There, the authors compare the user behavior with a goal model and ask for feedback when users achieve sub-goals or when they deviate from an expected sub-goal. This is similar to our detection of situations that might need an explanation.

Other work has focused on rationalizing and verbalizing the behavior of autonomous agents. Rationalizations do not need to accurately reflect the true decision-making process, but give some explanations like humans would give in similar situations. In~\cite{Ehsan2017} an agent's actions are rationalized by using an encoder-decoder neural network to translate between state-action information and natural language.
In \cite{RobotVerbalization} the agent's experiences on a route are verbalized by converting sensor data into natural language as answer to user queries with varying levels of abstraction, specificity and locality. 
Another approach to generate explanations at run-time is to use a multi-modal agent that can be queried `on-demand'~\cite{RobbICMI18, ChiyahINLG18}. There, the system behaviors are mapped into a modified version of fault trees, which the authors call \textit{model of autonomy}, that capture the possible states of the system~\cite{ChiyahHRI18}. The authors found that the explanations given by the agent helped improving the fidelity of the operators’ mental model, increasing the operator’s understanding of what the autonomous vehicles were doing and why, as well as how they work \cite{ChiyahINLG18}.
\section{The MAB-EX Loop for Explainability}\label{sec:mabex-loop}

\begin{figure}
    \centering
    \includegraphics[scale=0.6]{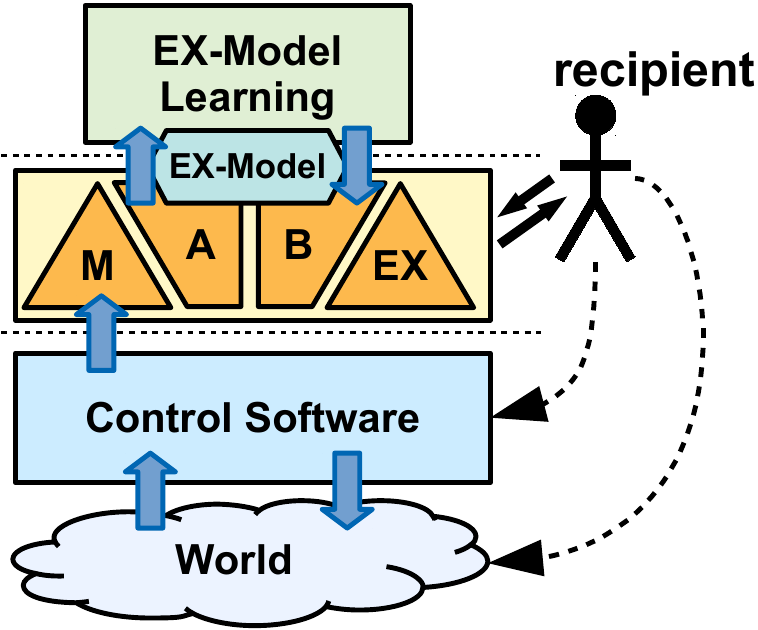}
    \caption{The \acf*{MAB-EX} framework.}
    \label{fig:mabex}
\end{figure}

Our framework for self-explaining systems is inspired by the \ac{MAPE} loop for self-adaptive systems, as we need similar self-reflection capabilities to detect the possible need for an explanation and to provide context-specific explanations.   
To achieve this, we propose the \acf{MAB-EX} framework as depicted in \cref{fig:mabex}. Note that the underlying system does not need to be self-adaptive. Our MAB-EX loop can be added to any kind of computing system. However, if the underlying system is self-adaptive, our approach can also be integrated into the existing (MAPE) feedback loop. Similar to the \ac{MAPE} loop, we first \emph{\textbf{M}onitor} the control system, its environment and possibly also the recipient of explanations. To this end, we capture and sample relevant sensor data,
(a history of) commands from controller components, and possibly also a history of user and/or system interactions and former explanations. To identify whether the user is satisfied with an explanation, we could also monitor the users face expressions (cf.~\cite{Cetal2001}).

Then, we \emph{\textbf{A}nalyze} the monitored data to detect an explanation need. This need can either be triggered because a recipient requires it (e.g., \enquote{Why are we leaving the highway?}) or because the system shows behavior that requires an explanation (e.g., \enquote{We are slowing down soon, because the road ahead is in poor condition.}). The latter can be detected by identifying 
deviations from formerly observed behavior that might indicate an explanation need. Examples are irregularities in the monitored sensor data or sudden changes in the way the user interacts with the system. In the former case we additionally need to analyze whether the change can be expected, e.g., due to a user interaction. Furthermore, the history of controller commands or user commands can be analyzed to identify aimless sequences of commands/ interactions (e.g., contradicting commands over time that lead to nowhere).
In case of explanation queries from the recipient, the query can be processed in this phase.

Instead of planning new behavior like in the MAPE loop, our third phase is to \emph{\textbf{B}uild}
an explanation by evaluating an internal model of the system, which we call \emph{\emodel}, based on the currently monitored system behavior, in order to extract relevant information. 
An \emodel{} is a behavioral model of the system that captures causal relationships between events and system reactions. It allows for identifying possible causes for the behavior that needs to be explained, e.g., traces of events that may lead to the behavior. It may also allow for look-ahead simulation to enable answering questions like \enquote{What happens if ... ?} or \enquote{When will ... be possible again?}. Possible implementations for an \emodel{} could, e.g., be (fault/decision) trees that connect observations to possible reasons, or executable behavior models (e.g., state machines), as illustrated in our case study in \cref{sec:case-study}.
Such models may be constructed from requirements or from a behavior model, constructed manually, or learned from observations. 
Possible implementations also could be 
goal models that capture goals, objectives and  motivations for the systems' behavior.
Note that this synthesized explanation is not yet in a recipient-understandable format, but in an intermediate format. With \emph{recipient} we refer to the addressee of an explanation, which can be a user (e.g., engineer or end-user) or a (sub-)system.

Thus, the fourth and last phase is to actually \emph{\textbf{EX}plain} the behavior in question to the recipient, meaning to transfer the result of the building phase to an understandable explanation for the target group. The explanation should be target-specific, as, e.g., an engineer might need more detailed information than an end-user, and an  end-user  might not understand  technical  terms  that  are useful for  the  engineer. To this end, we use a \emph{\tmodel}, e.g., mental model of a human recipient or an explanation interface between control software of  different  systems (e.g., to allow  for  collaborative learning and operation).
It describes preferences of the recipient w.r.t. explanation format (e.g., textual, image, voice, or machine-processable) and kind of information that should be included in an explanation (e.g., level of abstraction, points of interest). These \tmodel{}s can range from general mental models for target groups (e.g., engineers vs.\ end-users) to models for individual users.\\
The final explanation is provided to the recipient and thus, we do not have a loop anymore. However, we might have an indirect loop by monitoring the recipient's reaction to the given explanation or because the recipient itself asks for more details on the explanation.

As both, the system that needs to be explained and the recipient of the explanation may evolve over time or are subject to uncertainties at design time (about the system behavior, its operational context, and the recipient and its preferences), we include a \emph{\textbf{Model Learning}} into our framework that is responsible for updating both our \emodel{} and our \tmodel.
Consider, e.g., the case that an emergency maneuver is executed due to a spontaneously changing extreme weather condition for autonomous driving. If the monitored and analyzed behavior is not contained in the \emodel, an explanation cannot be built immediately in the building phase. However, after having integrated this new behavior into the \emodel{}, an explanation can be provided later or if the behavior should occur again. 
Model Learning can be realized using machine learning algorithms, or as an expert system, where domain experts (and probably also other cooperating systems) are asked to provide an explanation for the observed situation, or as combination of both. This cooperative updating process could, e.g., be realized by connecting \emph{Model Learning} components of different systems and experts via a cloud service. 
To update the \tmodel{}, preferences of the recipient can be inferred from the interaction with the recipient (e.g., based on follow-up questions that indicate the wish for further information).

\section{Example Realizations of \ac{MAB-EX}}
\label{sec:case-study}

\newcommand{\smalltextsf}[1]{\textsf{\small#1}}
\newcommand{\SCENARIOCarRegistersAtObstacle}{\smalltextsf{Car\-Registers\-At\-Obstacle}}
\newcommand{\SCENARIOCarEnteringAllowedDefault}{\smalltextsf{Car\-Entering\-Allowed\-Default}}
\newcommand{\SCENARIOCarEnteringDisallowedWhenCarPassing}{\smalltextsf{Car\-Entering\-Disallowed\-When\-Car\-Passing}}
\newcommand{\SCENARIOEnteringDisallowedForOtherPriorityVehicle}{\smalltextsf{Entering\-Disallowed\-For\-Other\-Priority\-Vehicle}}
\newcommand{\SCENARIOSetPriorityForEmergencyVehicle}{\smalltextsf{Set\-Priority\-For\-Emergency\-Vehicle}}

\begin{figure}
	\centering
	\includegraphics[width=\columnwidth]{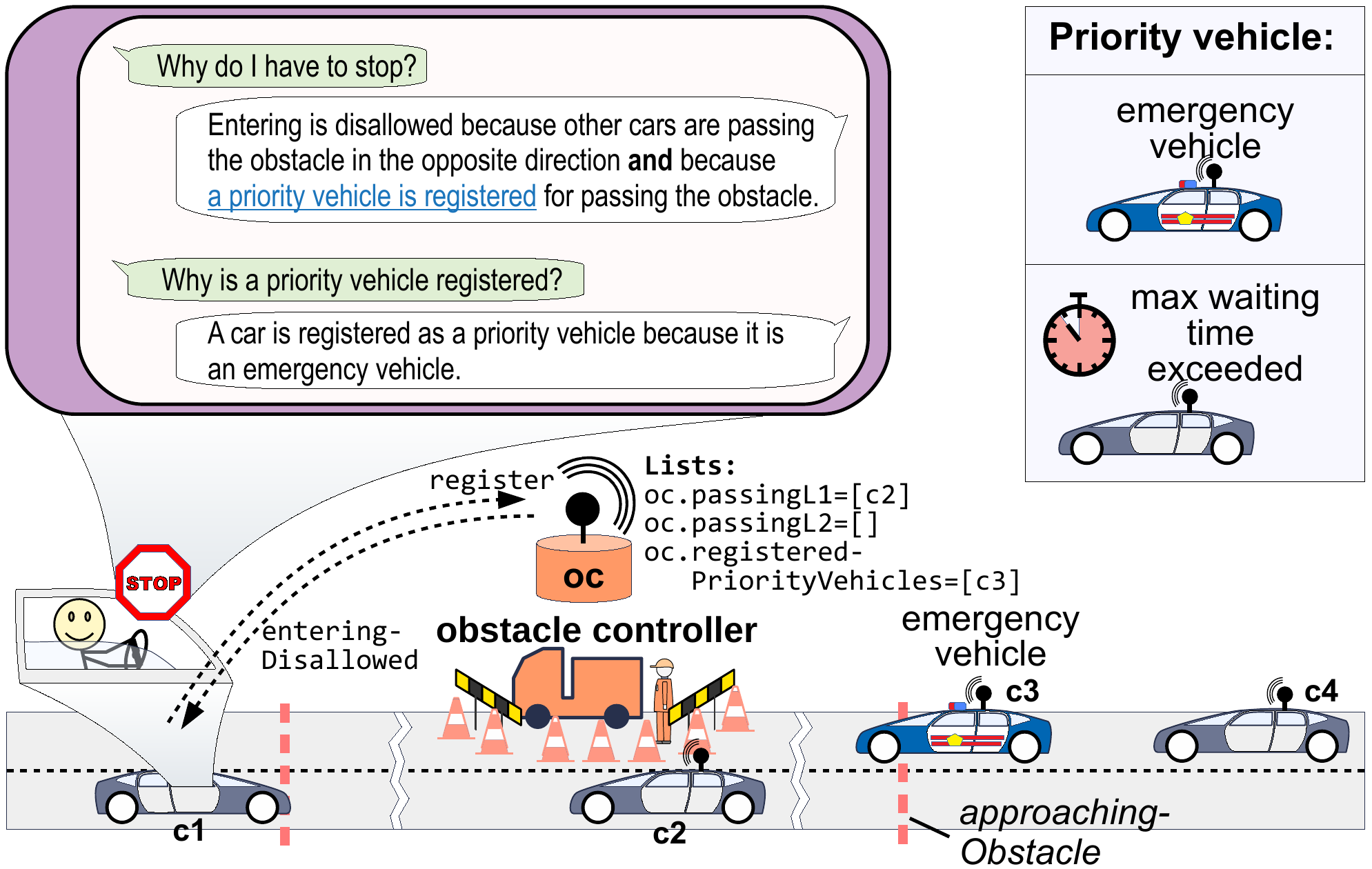}
	\caption{\acf*{V2X} narrow passage coordination assistance system}
	\label{fig:car-to-x-example-sketch}
\end{figure}

We illustrate the \ac{MAB-EX} framework by instantiating it for an example of an advanced \ac{V2X} driver assistance system such as is typically envisioned for future cooperative mobile systems~\cite{sommer2014vehicular}. This system helps drivers safely pass obstacles on the road (see \cref{fig:car-to-x-example-sketch}). In this example, cars that approach the obstacle register at an obstacle controller and await permission to enter the narrow street section. The system's response (pass or stop) is displayed to the driver. We focus on a car (\textsf{c1} in \cref{fig:car-to-x-example-sketch}) that must stop and where the driver wonders why passing the obstacle is not possible---even though the roadworks is on the opposite lane and the road ahead seems free.
We envision that an interface (top left of \cref{fig:car-to-x-example-sketch}) provides an explanation to this question. The explanation in this case is twofold: There is a car in the narrow street section approaching from the other side (which the driver may not see yet), and, moreover, there is an emergency vehicle approaching from the other side, which is not yet in the narrow section, but has registered at the obstacle controller as a priority vehicle. Other reasons to stop could be fairness to cars that already waited for a long time.

We illustrate the four building blocks of the framework for this example.

\subsection{Monitor} 
As stated in \Cref{sec:mabex-loop}, we monitor the controlled system to identify situations that demand an explanation. In the example, we need information about the position of a car in the lane (\smalltextsf{L1} or \smalltextsf{L2}) and the controller's response towards the event of approaching the obstacle (\smalltextsf{enteringDisallowed} or \smalltextsf{enteringAllowed}).
Since we are only interested in one specific situation in this example, we 
do not 
need more information. In extended scenarios, it may be interesting to monitor, for example, the vehicle's speed to identify critical situations that may demand an explanation. If the system has a query feature for explanations in its HMI, we need to monitor user queries as well.  

\subsection{Analyze} 
We analyze the monitored data and identify situations that need to be explained. In our example, the only situation that needs an explanation is when a car is approaching the obstacle on lane \smalltextsf{L1} and the controller responds with \smalltextsf{enteringDisallowed}. 

\subsection{Build} 
Building the actual explanation is usually the 
greatest challenge
in the MAB-EX framework. We present two solutions that can be used to identify and compose the ingredients of an explanation (i.e., the causes of the event that needs to be explained). The first solution to provide such explanations is based on \textit{models of causality}, which connect actions of the system to their (possibly internal) causes, including natural language descriptions. The second solution shows how behavioral models of the system that are created at design-time can be leveraged to assemble explanations at run-time.    

\subsubsection{Models of Causality Approach}
This approach was explored before for providing explanations to operators of autonomous underwater vehicles~\cite{ChiyahHRI18, ChiyahINLG18}. We can also apply this approach to the example above.

\begin{figure}
	\centering
	\includegraphics[width=\columnwidth]{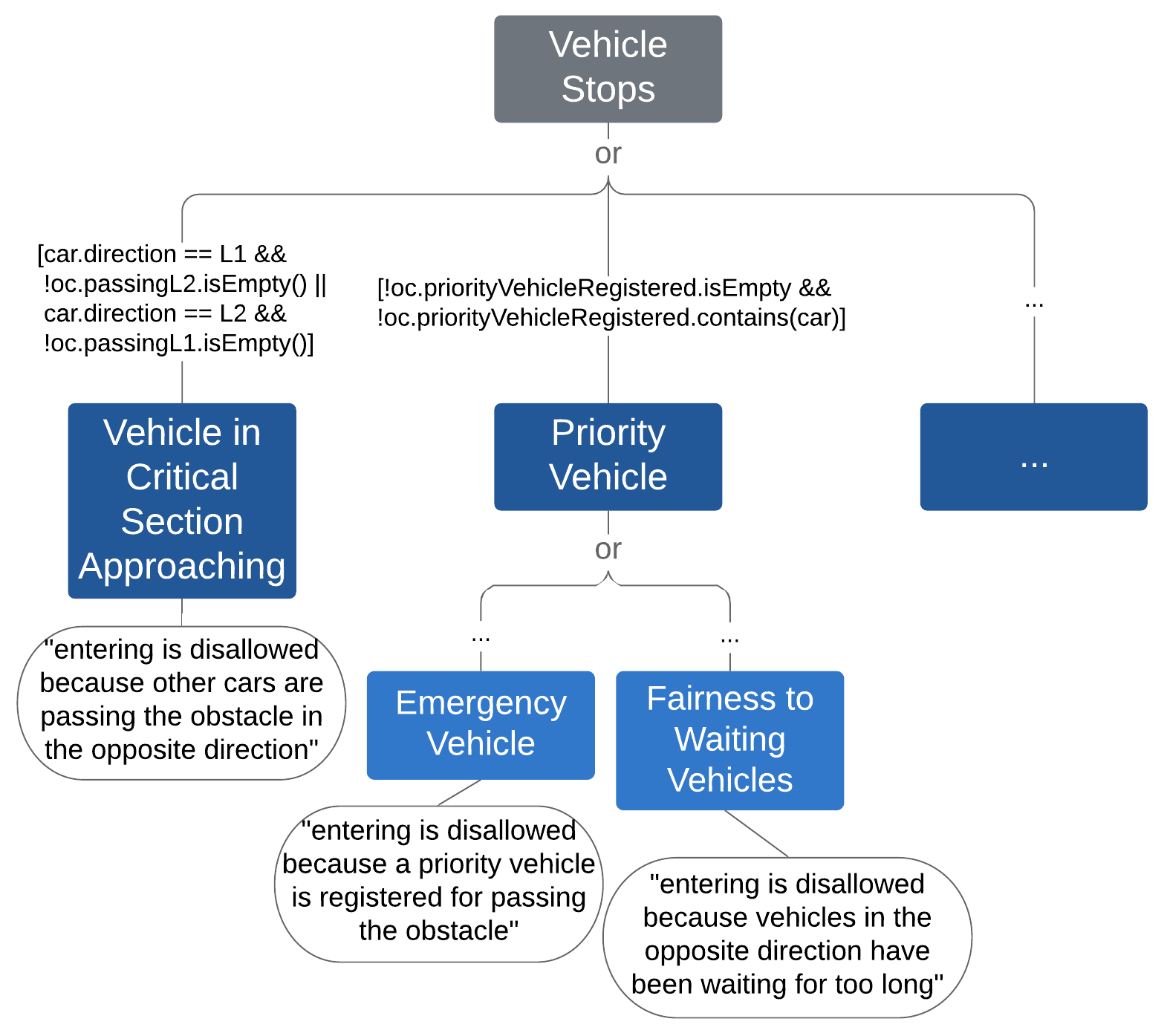}
	\caption{Model of causality for the car example. Each node has a condition of system variables and a natural language explanation. A model of causality may contain as many nodes and depth levels as necessary.}
	\label{fig:car-example-tree}
\end{figure}

\Cref{fig:car-example-tree} shows a tree for the situation of when a car stops in our case example. The root node is the observable event (\enquote{Vehicle Stops}) and the branches give the possible reasons for the event. Traversing down the tree gives the explanations, which are attached to the nodes as natural language sentences. The explanations are increasingly detailed further down the tree, allowing to easily adapt to the user's needs. Together with the explanations, the nodes have a condition in terms of system variables that can be checked to determine if the node could be a plausible reason for the observable event. 

A solution based on models of causality gives a high-level representation of the events without looking at the system details. Thus, the trees are independent from the system implementation, which allows to build these trees at any point of the system's life. They can be directly derived at the requirement specification phase or built after the system has been released. 
This approach requires minimal or no changes to the system that it explains, as they only monitor system variables to evaluate the node conditions. 
The level of abstraction over the system's internal processes makes controlling the amount of information disclosed easily adjustable. 

However, the models of causality rely on manual modeling, which involves system knowledge that only those building or maintaining the system itself can provide. They also require knowing ahead of time which events can happen and the different explanations for the phenomena. Thus, they are limited (or focused) to explaining certain predefined state conditions.









\subsubsection{Creating Explanations Dynamically from Run-Time Models}

The above approach has the advantage that designers can easily model the system's explanation capabilities, specifically control the level of abstraction of the explanations, and that it can be easily integrated into a system: 
However, the explanations are limited to the phenomena anticipated at design time, and they are limited to explanations concerning specific (current or past) states---properties of sequences of states or predictions about the future are not possible. To achieve this, we require more elaborate models at run-time, which connect the behavior of the system and its environment with requirements specification, assumption specifications, and which can be queried and, especially, executed for look-ahead predictions.

Such an approach could be based on executable scenario-based behavior models, e.g., \ac{LSC}~\cite{Damm2001}, that can be annotated with per-requirement/scenario rationales, or could contain trace links to natural language requirements, from which explanations can then be derived dynamically, at run-time. The executable scenarios could live solely within the system's explanation layer (as EX models), but they can even be used as the final implementation code for the distributed reactive behavior of CPS such as our example V2X system~\cite{Greenyer2018d}.

We sketch a scenario-based explanation approach in the following. \Cref{list:sml-specification} shows scenarios from the example \ac{V2X} system in \ac{SML}~\cite{Greenyer2017a}, a textual language for modeling \ac{LSC}-style scenarios. An SML specification models, via  \textit{assumption-} and \textit{guarantee scenarios}, how objects of a system and its environment interact by sending messages. Guarantee scenarios describe how the developed (software) system may, must, or must not react to environment events; assumption scenarios (not shown here) describe what may, will, or will not happen in the environment. Each scenario models valid sequences of events, using different \textit{modalities}. For example, events can be \textit{requested}, which means that the event must eventually occur; non-requested messages need never occur. Events can also be \textit{strict}, saying that when the scenario is waiting for the event to occur, no event must occur that is expected within the same scenario at an earlier or later point. The \textit{forbidden} modality models events that are forbidden while (a certain part of) a scenario is active; \textit{interrupt} models events that are allowed, but will interrupt the scenario. The scenarios are \textit{executable}; at execution-time, multiple scenarios can be active at the same time, each requesting or forbidding certain events, and events are chosen to satisfy all the constraints imposed by the scenarios.

	\begin{lstlisting}[caption=\acs*{SML} scenarios for the \acs*{V2X} driver assistance system,
label=list:sml-specification,
style=SMLXStyle,
float,
floatplacement=H
t]
...
guarantee scenario CarRegistersAtObstacle
	bindings [oc = cp.obstacleCtrl] {
	sensor -> car.approachingObstacle()
	//@EX: when approaching an obstacle, the car must register at the obstacle control 
	strict requested car -> oc.register()
}

guarantee scenario CarEnteringAllowedDefault {
	car -> oc.register()
	// @EX: entering is allowed because there is no indication to disallow it.
		requested oc -> car.enteringAllowed() 
} constraints [
	interrupt oc -> car.enteringDisallowed()
]

guarantee scenario CarEnteringDisallowedWhenCarPassing {
	car -> oc.register()
	alternative [car.direction == L1 && !oc.passingL2.isEmpty() || car.direction == L2 && !oc.passingL1.isEmpty()] {
		// @EX: entering is disallowed because other cars are passing the obstacle in the opposite direction.
		strict requested oc -> car.enteringDisallowed()
	} constraints [
		forbidden oc -> car.enteringAllowed()
	]
}

guarantee scenario EnteringDisallowedForOtherPriorityVehicle {
	car -> oc.register()
	alternative [!oc.registeredPriorityVehicles.isEmpty()
			&& !oc.registeredPriorityVehicles.contains(car)]{
		// @EX: entering is disallowed because a priority vehicle is registered for passing the obstacle.
		strict requested oc -> car.enteringDisallowed()
	} constraints [
		forbidden oc -> car.enteringAllowed()
	]
	
guarantee scenario SetPriorityForEmergencyVehicle {
	car -> oc.register()
	alternative [car instanceOf EmergencyVehicle] {
		// @EX: car registered is a priority vehicle because it is an emergency vehicle.
		strict committed oc -> oc.registeredPriorityVehicles.add(car)
	}
}	
}
...
\end{lstlisting}

The scenario \SCENARIOCarRegistersAtObstacle{} specifies that when a car sensor detects that the car approaches an obstacle, the car must register at the obstacle controller \smalltextsf{oc}. The scenario \SCENARIOCarEnteringAllowedDefault{} specifies that the obstacle control shall allow the car to enter, unless the scenario is interrupted by the \smalltextsf{entering\-Disallowed} event that can be requested, for example, by the scenario \SCENARIOCarEnteringDisallowedWhenCarPassing{}, which models the case of a car that is passing the obstacle in the other direction. Scenario \SCENARIOEnteringDisallowedForOtherPriorityVehicle{} models the case where a priority vehicle is registered for passing the obstacle while the car that is subject to that scenario is itself not a priority vehicle. Last, \SCENARIOSetPriorityForEmergencyVehicle{} specifies that when an emergency vehicle registers at the obstacle control, it will be added to the list of \smalltextsf{registeredPriorityVehicles} (cf.\,\cref{fig:car-to-x-example-sketch}).

\begin{figure}
	\centering
	\includegraphics[width=\columnwidth]{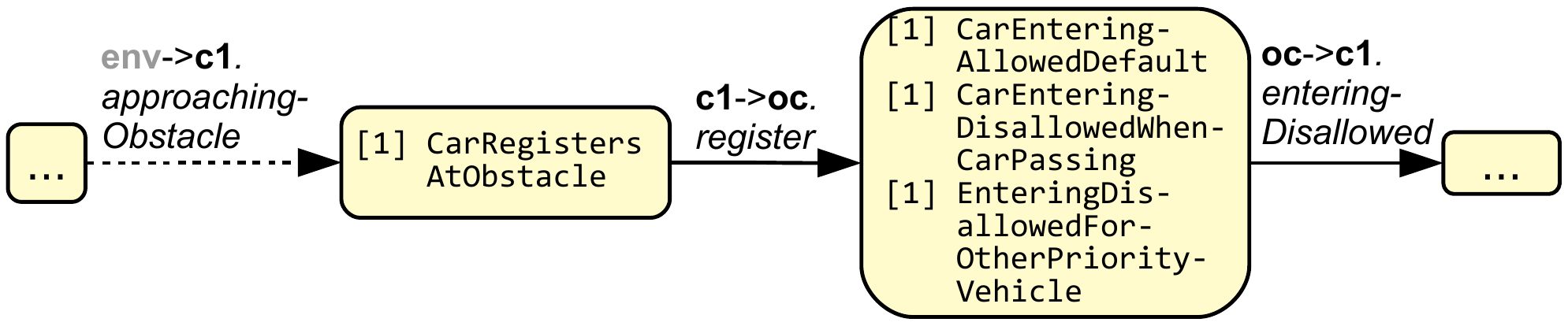}
	\caption{Scenario run-time states for the V2X example}
	\label{fig:car-to-x-scenario-runtime-states}
\end{figure}

\Cref{fig:car-to-x-scenario-runtime-states} shows a sequence of states in the execution of these scenarios. For brevity we omit the states of the underlying objects. Starting with the \smalltextsf{approaching\-Obstacle} event the scenario \SCENARIOCarRegistersAtObstacle{} is activated. Then,  \smalltextsf{register} terminates \SCENARIOCarRegistersAtObstacle{}, but activates \SCENARIOCarEnteringAllowedDefault{}, \SCENARIOCarEnteringDisallowedWhenCarPassing{}, and \SCENARIOEnteringDisallowedForOtherPriorityVehicle{}. In this case, \smalltextsf{enteringDisallowed} is executed due to the conditions in the latter two scenarios that are satisfied in the state as in \cref{fig:car-to-x-example-sketch}.

The scenarios contain explanations annotated to all events that they request. This way, an explanation as show in \cref{fig:car-to-x-example-sketch} can be produced by combining these explanations. \Cref{fig:car-to-x-example-sketch} also shows that the explanation component is able to answer the follow-up question \textit{Why is a priority vehicle registered?}. This question can be answered by traversing over the past states in search of the events that contributed to rendering the condition true. In this example, a past activation of \SCENARIOSetPriorityForEmergencyVehicle{} triggered by the \smalltextsf{register} message from the emergency vehicle \smalltextsf{c3} caused the event of adding \smalltextsf{c3} to the list of \smalltextsf{registeredPriorityVehicles}, which was the point from when the evaluation of the condition in \SCENARIOEnteringDisallowedForOtherPriorityVehicle{} turned from \textit{false} to \textit{true}. 

The scenarios could also be used for a forward-exploration of possible future behaviors that could be used to answer questions about the future, such as \textit{When will I be allowed to pass the obstacle?} Moreover, instead of annotating the scenarios with explanations, these could also be extracted from textual requirements that could be referenced via trace links. It will be interesting to elaborate how also explanations for not executing certain events can be provided.


\subsection{Explain} 
Finally, we need to generate the actual explanation from the information gathered in the \emph{Build} component. In our example, the explanation is given as a text that is generated from the text fragments associated with the nodes in the model of causality or with the annotations in the scenario specifications. This way, the provided explanation for the detected situation is rendered as \emph{``Entering is disallowed because other cars are passing the obstacle in the opposite direction and a priority vehicle is registered for passing the obstacle``}.






\section{Conclusion and Research Roadmap}\label{sec:discussion}

The MAB-EX approach towards explainability of system run-time behavior
represents a first approach towards a generalizable architecture for self-explainable systems. 
We have shown how requirements- and models-at-runtime can be exploited as a basis for realizing self-explanation capabilities.



The road towards truly comprehensible, flexibly tailored explanations yields many challenges:
  \absatz{Comprehensible explanations} Useful explanations demand for a representation of decisions that supports tailoring the abstraction of explanation parts to the recipient, e.g., in contrast to the infamous Windows operating system blue screens 'explaining' its failure in terms of memory locations. Similarly, in engineering CPS with domain experts, a networking engineer might be very interested in communication decisions, but less in HCI decisions the system has made. 
  \absatz{Explanation presentation} Depending on the facts to be explained or the receivers' background, different presentations of explanations will be of different usefulness. While engineers might prefer textual explanations (e.g., log files), users might prefer graphical explanations or conversational interfaces.
  \absatz{Focused explanations} To prevent systems from overwhelming receivers with potentially relevant information we need to conceive means for filtering and truncating explanation information based on, e.g., user studies or learned patterns.
  \absatz{Consultable explainers} When systems are capable of producing a wealth of explanations of different extent, abstraction, and personalization, being able to consult systems for specific explanations becomes necessary to support producing the best-possible explanations for different circumstances.
  \absatz{Interactive explanations} Similar to human discourse, self-explaining systems may produce explanations that entail subsequent queries about the reasons for a given explanation. Consequently, truly useful self-explaining systems should support interactive exploration of explanations, explanation sequences, and metadata (e.g., relations between explanations).
  \absatz{Explanation prediction} Systems equipped with means for self-explanation should be able to explain the future potential behavior as well as why expected events did not happen. This could have the form of explicit what-if queries or online explanation about expected behavior.
  \absatz{Cooperative explanations} To understand the behavior of systems cooperating in the Internet of Things, the smart factory of the future, or in \ac{V2X}, systems must be able cooperatively explain their behavior. This demands for means to align their explanation terminologies (e.g., through explanation ontologies for specific domains) and might require reason about their own behavior based on implicit explanations (i.e., observations) of the cooperating systems' behaviors. 
  \absatz{A posteriori explaining} The long-lived systems in industrial domains will need to cooperate with systems incapable of explaining themselves. Therefore, means to explain system behavior based on observations made by a posteriori deployed, dedicated explainers is necessary.


\bibliographystyle{IEEEtran}
\bibliography{references}

\end{document}